\documentclass[10pt,twocolumn,letterpaper]{article}
\usepackage[pagenumbers]{cvpr}
\usepackage[T1]{fontenc}
\usepackage{amsmath,amssymb}
\usepackage{graphicx}
\usepackage{booktabs}
\usepackage[expansion=false]{microtype}
\DeclareMathOperator*{\argmin}{arg\,min}

\makeatletter
\patchcmd{\@maketitle}{\null}{\null\vspace{-5em}}{}{}
\makeatother
\definecolor{cvprblue}{rgb}{0.21,0.49,0.74}
\usepackage[pagebackref,breaklinks,colorlinks,allcolors=cvprblue]{hyperref}
\def\confName{CVPR}
\def\confYear{2027}
\title{RoboMorpher: Scaling Robot Morphology Representations for Cross-Embodiment Action Prediction}
\author{%
Wenwei Lin$^{1,*}$, Yan Qin$^{2,*}$, Weiliang Deng$^{3,*}$, Zixuan Li$^{1,*}$, Kaixuan Wang$^{1,*}$,\\
Zanxin Chen$^{4}$, Yue Chen$^{5}$, Tianshuo Yang$^{1}$, Junting Chen$^{6}$, Zhixuan Liang$^{1}$,\\
Yao Mu$^{4}$, Ximin Lyu$^{3}$, Renjing Xu$^{2}$, Ping Luo$^{1,\dagger}$, Tianxing Chen$^{1,\dagger}$\\[5pt]
{\small $^{1}$The University of Hong Kong \quad $^{2}$HKUST (Guangzhou) \quad $^{3}$Sun Yat-sen University}\\
{\small $^{4}$Shanghai Jiao Tong University \quad $^{5}$Peking University \quad $^{6}$National University of Singapore}\\[3pt]
{\small $^{*}$Equal contribution \qquad $^{\dagger}$Corresponding authors}\\[5pt]
{\normalsize\textbf{Project page:} \href{https://robomorpher.github.io/}{\texttt{https://robomorpher.github.io/}}}
}
\hypersetup{
    pdftitle={RoboMorpher: Scaling Robot Morphology Representations for Cross-Embodiment Action Prediction},
    pdfauthor={Wenwei Lin, Yan Qin, Weiliang Deng, Zixuan Li, Kaixuan Wang, Zanxin Chen, Yue Chen, Tianshuo Yang, Junting Chen, Zhixuan Liang, Yao Mu, Ximin Lyu, Renjing Xu, Ping Luo, Tianxing Chen}
}

\begin{document}
\maketitle
\begin{abstract}
Learning shared action representations is central to cross-embodiment manipulation. However, identical joint values can produce different spatial configurations across robots, limiting representations learned from joint values alone. We introduce RoboMorpher, a morphology-conditioned action tokenizer that jointly reconstructs joint angles and link poses to associate control coordinates with their kinematic effects. MorphGen supplies scalable kinematic supervision by generating serial-arm descriptions from survey-derived geometric priors. We pretrain RoboMorpher on 40 million morphology--state pairs and integrate the frozen tokenizer into a vision--language--action policy for discrete prediction and continuous refinement. We also introduce EmbodiBench, a ten-task benchmark with 30 generated URDFs per task. With downstream URDFs excluded from tokenizer pretraining, the policy achieves 56.40\% success on RoboTwin~2.0 and 23.27\% on EmbodiBench, exceeding continuous joint-space control by 9.73 and 15.34 percentage points. Removing link-pose supervision reduces EmbodiBench success by 17.94 percentage points, supporting morphology-aware geometric supervision for joint-action representation learning.
\end{abstract}

% Queue the survey figure on page 1 for the top of page 2.
\begin{figure*}[t]
    \centering
    % Equal height; width allocation follows the source PDF aspect ratios.
\begin{tabular}{@{}l@{\hspace{0.02\textwidth}}l@{}}
{\small\textbf{(a) Joint-wise length priors}} & {\small\textbf{(b) MorphGen samples}}\\[4pt]
\includegraphics[height=0.3747058824\textwidth]{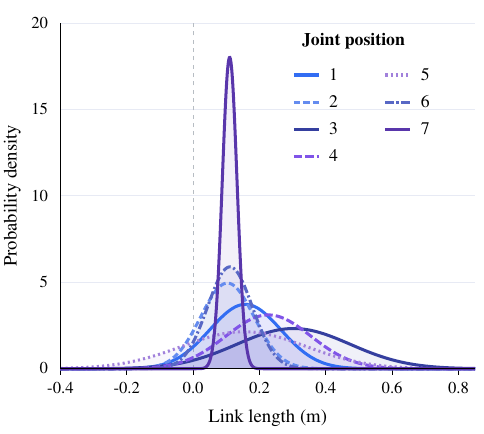} &
\includegraphics[height=0.3747058824\textwidth]{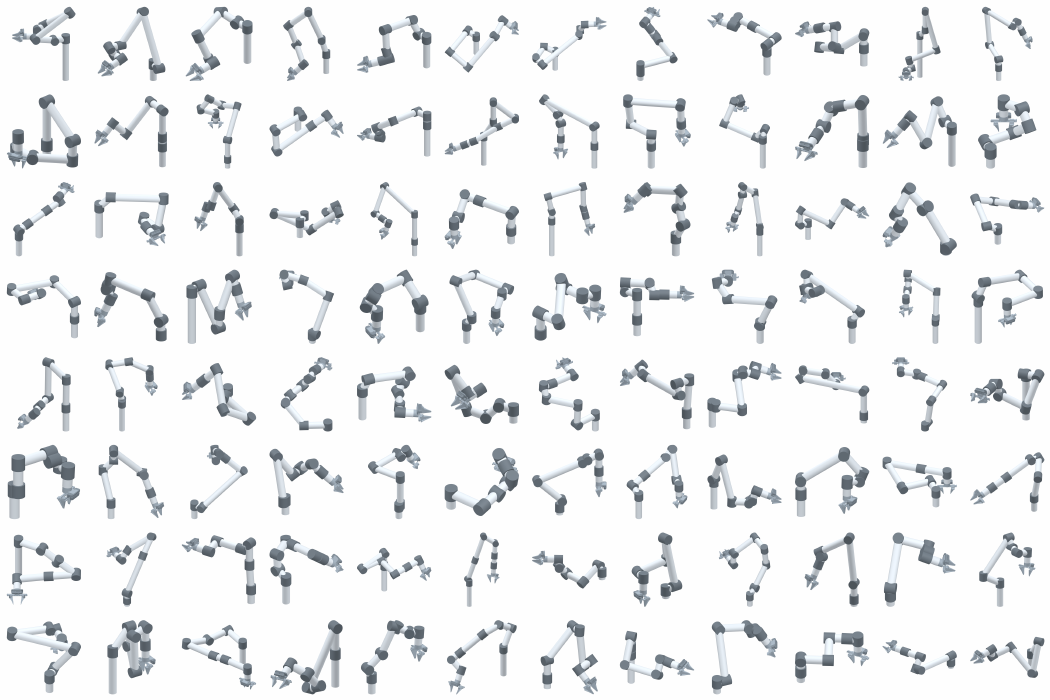}
\end{tabular}

    \caption{\textbf{Survey and morphology generation.} (a) Seven Gaussian priors $\mathcal{N}(\mu_i,\sigma_i^2)$ fitted to 185 reported lengths from 30 description records. Curves show the fitted distributions before truncation and chain-length rescaling. (b) Ninety-six generated morphologies varying link geometry and joint-frame arrangement. \Cref{app:survey,app:morphgen} provide the survey records and generation details.}
    \label{fig:morphgen-gallery}
\end{figure*}

\section{Introduction}
\label{sec:intro}

Generalist robot policies learn from demonstrations across platforms~\cite{oxe,octo,rdt}, but the kinematic meaning of joint-space actions depends on robot morphology. Identical joint values can produce different link poses as link lengths and joint arrangements vary; padding actions to a common dimension does not resolve this ambiguity. End-effector (EE) control provides Cartesian targets but requires robot-specific inverse kinematics and does not uniquely specify the full joint configuration. A shared joint-action representation should therefore preserve control coordinates while accounting for their morphology-dependent spatial effects.

To learn this relationship without collecting manipulation demonstrations for every morphology, we use robot kinematics as supervision. We introduce \emph{MorphGen}, a procedural generator informed by a survey of 33 URDF/MJCF descriptions. MorphGen normalizes serial-arm kinematics into a common local-$Z$ convention, samples six- and seven-joint manipulators from empirical link-length priors, and selects feasible home poses through collision-aware inverse kinematics. Forward kinematics provides link-pose targets for 40 million morphology--state pairs.

We present \emph{RoboMorpher}, a morphology-conditioned action tokenizer pretrained on these data. Morphology conditions its encoder and decoder, while kinematic-chain attention biases encode relative joint positions along the chain. Joint-angle reconstruction preserves control coordinates, and link-pose reconstruction supervises their kinematic effects. The pretrained tokenizer is then frozen and integrated into a vision--language--action (VLA) policy. Following Discrete Forcing~\cite{discreteforcing}, a discrete branch predicts action codes, which are decoded into coarse trajectories and refined by a continuous flow-matching branch. Pretrained morphology features condition both branches.

We evaluate on 15 RoboTwin~2.0 tasks across five embodiments~\cite{chen2025robotwin} and introduce \emph{EmbodiBench}, comprising ten tabletop tasks with 30 generated URDFs each. Downstream URDFs are excluded from tokenizer pretraining; policies learn from task demonstrations. RoboMorpher achieves 56.40\% and 23.27\% success, exceeding continuous joint-space control by 9.73 and 15.34 percentage points. Removing link-pose supervision reduces EmbodiBench success to 5.33\%, supporting its contribution to downstream performance.

Our contributions are as follows.
\begin{itemize}
    \item \textbf{MorphGen}, a survey-guided generator of robot morphologies and large-scale kinematic pretraining data.
    \item \textbf{RoboMorpher}, a morphology-conditioned action tokenizer trained through joint-angle and link-pose reconstruction, integrated into a VLA policy for discrete prediction and continuous refinement.
    \item \textbf{EmbodiBench}, a benchmark for manipulation across generated morphologies, with experiments on task performance, representation design, and reconstruction fidelity.
\end{itemize}

\section{Related Work}
\label{sec:related}

\paragraph{Cross-Embodiment Robot Learning.}
Open X-Embodiment aggregates demonstrations across robot platforms~\cite{oxe}, supporting generalist policies such as Octo and RDT~\cite{octo,rdt}. HPT and CrossFormer learn shared representations from heterogeneous embodiments~\cite{wang2024hpt,doshi2025crossformer}. MetaMorph and Shared Modular Policies incorporate robot morphology through policy conditioning and inter-module communication~\cite{metamorph,smp}, while Body Transformer and BodyVLA exploit kinematic structure in policy architectures~\cite{pmlr-v270-sferrazza25a,suzuki2026embedding}. UniAct, X-VLA, and XL-VLA further explore universal actions, embodiment-specific prompts, and shared latent action spaces~\cite{zheng2025uniact,zheng2025xvla,Jiang_2026_CVPR}. Our focus is morphology-conditioned action encoding and decoding, learned from kinematic supervision before policy training.

\paragraph{Robot Action Representations and Tokenization.}
Vector quantization learns discrete representations through reconstruction~\cite{vqvae}. FAST, VQ-VLA, and FASTer explore action tokenization for efficient sequence modeling~\cite{fast,wang25vqvla,liu2025faster}, while X-Tokenizer aligns discrete action codes with pretrained vision-language representations~\cite{kang2026xtokenizer}. Discrete Forcing combines discrete action prediction with continuous flow-matching refinement~\cite{discreteforcing}. RoboMorpher extends configuration-level tokenization with morphology conditioning and joint-angle/link-pose reconstruction, explicitly supervising the relationship between control values and spatial effects.

\paragraph{Robot Data Generation and Benchmarks.}
RoboGen synthesizes robot tasks, scenes, and training supervision through simulation~\cite{robogen}. RoboTwin~2.0 and RoboDojo provide diverse manipulation tasks and evaluation across simulation and real-world settings~\cite{chen2025robotwin,chen2026robodojo}. MorphGen targets kinematic supervision by generating chains and configuration--pose pairs from survey-derived priors. EmbodiBench evaluates the resulting action representations across generated morphologies on ten tabletop tasks.

\section{Method}
\label{sec:method-overview}
\begin{figure*}[t]
    \centering
    \includegraphics[width=\textwidth]{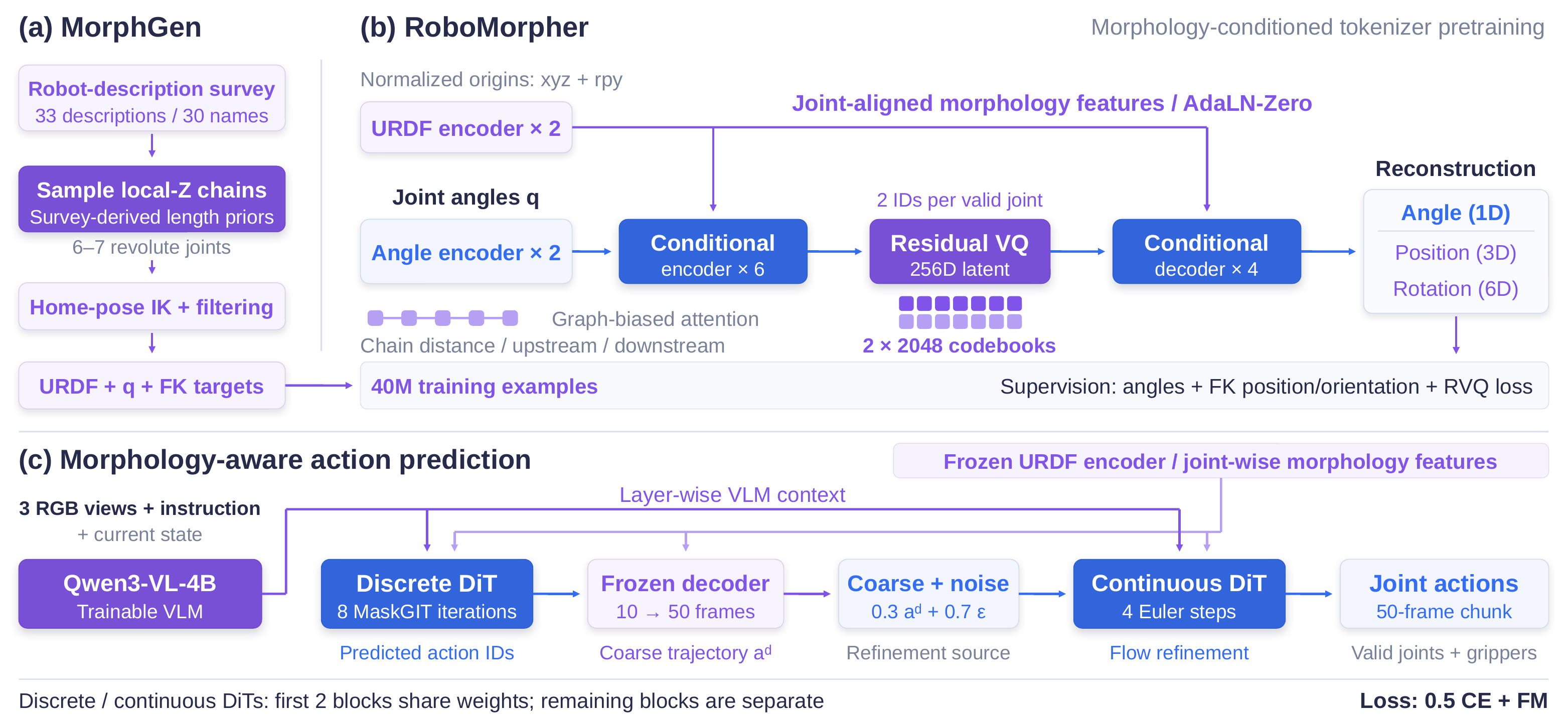}
    \caption{\textbf{RoboMorpher pipeline.} (a) MorphGen samples local-$Z$ chains using survey priors and IK filtering. (b) URDF-conditioned tokenization learns two codes per joint with FK supervision. (c) The VLA predicts codes, decodes ten nodes, interpolates 50 frames, and refines them with flow matching. Frozen URDF features condition both DiTs and the decoder. VLM context conditions the DiTs.}
    \label{fig:architecture}
\end{figure*}

Our pipeline separates kinematic representation learning from task-specific action selection (\cref{fig:architecture}). MorphGen generates morphology--configuration pairs with forward-kinematics (FK) supervision. RoboMorpher encodes configurations into joint-aligned codes and reconstructs angles and link poses conditioned on morphology. A VLA policy predicts these codes and refines decoded trajectories while keeping the tokenizer frozen.

\subsection{MorphGen Data Generation}
\label{sec:morphgen}
MorphGen converts survey-derived geometric priors into serial-arm descriptions, represents their kinematics in a common frame convention, and selects home poses using collision-aware inverse kinematics (IK). FK then supplies configuration--pose supervision without manipulation demonstrations.

\subsubsection{Robot Morphology Survey and Geometric Priors}
\label{sec:survey}
We survey 33 serial-manipulator descriptions covering 30 robot names. Gen3, Koch v1.1, and Panda each have URDF and MJCF entries. Six- and seven-DoF arms account for 30/33 descriptions (90.9\%), motivating our sampling range. Variation in joint limits and tool coupling motivates explicit arm-chain extraction and separate gripper handling. \Cref{tab:survey-summary,app:survey} summarize the inventory and description ambiguities.

% Generated by scripts/build_tables.py; edit the source/script.
\begin{table}[t]
\centering
\small
\caption{\textbf{Robot-description survey.} Entry-level statistics. Masses and reaches are reported values. The appendix documents description-level discrepancies.}
\label{tab:survey-summary}
\begin{tabular}{@{}lr@{}}
\toprule
Property & Count / range \\
\midrule
Description entries / distinct names & 33 / 30 \\
URDF / MJCF entries & 24 / 9 \\
Gripper / other tool / no tool listed & 11 / 1 / 21 \\
Approx. reach (32 entries) & 0.25--2.05 m \\
Model mass (32 entries) & 0.226--33.5 kg \\
\bottomrule
\end{tabular}
\end{table}

A separate set of 30 link-length records contains 185 reported values across seven joint positions (\cref{fig:morphgen-gallery}(a)). The positions have 30, 30, 30, 30, 30, 27, and 8 observations, respectively. Zeros are retained and missing entries excluded. Positions 3 and 4 have the largest means (0.3029 and 0.2270\,m), while position 5 has the largest sample standard deviation (0.1862\,m). These differences motivate position-specific Gaussian priors. The priors characterize the surveyed inventory, with only eight observations supporting the seventh position.

\subsubsection{A Unified Kinematic Representation}
\label{sec:urdf-representation}
Because local frames and intermediate fixed links can differ between equivalent descriptions, we express each ordered serial chain in a common local-$Z$ convention. Let $O_i\in\mathrm{SE}(3)$ be joint $i$'s origin transform after composing intervening fixed transforms, and $a_i$ its rotation axis. We choose $C_i$ such that $C_i e_z=a_i$, using the minimum-angle rotation and a fixed $X$-axis choice for the antiparallel case. With $C_0=I$,
\begin{equation}
    \widetilde O_i=C_{i-1}^{-1}O_iC_i,
    \qquad \widetilde T_i(q)=\prod_{k=1}^{i}\widetilde O_k R_z(q_k).
    \label{eq:localz}
\end{equation}
Rotations are embedded in homogeneous transforms, $R_z$ is a local-$Z$ rotation, and $T_i$ is the original root-to-link transform. The identity $\widetilde T_i(q)=T_i(q)C_i$ preserves joint values, signs, zeros, and physical joint positions. Inverse frame corrections recover link orientations. Asset-origin corrections preserve visual and collision placement.

The tokenizer receives $\mathcal M=\{(t_i,\phi_i)\}_{i=1}^{J}$, comprising each normalized origin's translation and XYZ roll--pitch--yaw angles, together with a validity mask. Joint angles are supplied separately. This six-value interface removes explicit axis parameters and intermediate fixed paths. It supports independent revolute/continuous chains and rejects selected prismatic/mimic joints or skipped movable joints. This standardizes the input convention without defining a unique canonical form for physical morphology.

\subsubsection{Survey-Guided Procedural Generation}
\label{sec:urdf-generation}
The generator samples six- or seven-joint revolute arms in the local-$Z$ convention. An initial length $\widetilde\ell_i$ is drawn from a truncated Gaussian using the mean $\mu_i$ and sample standard deviation $\sigma_i$ of survey column $i$ (\cref{tab:morphgen-prior}). The lower bound is zero and the upper bound is $\max(u,\mu_i+0.1)$\,m, where $u$ is 0.65\,m with probability 0.95 and 0.35\,m otherwise. The lengths are rescaled to a sampled total chain length,
\begin{equation}
    \ell_i=L\frac{\widetilde\ell_i}{\sum_k\widetilde\ell_k},\qquad
    L\sim\begin{cases}
    \mathcal U(0.8,1.6)&0.95,\\
    \mathcal U(1.6,2.0)&0.03,\\
    \mathcal U(0.3,0.8)&0.02.
    \end{cases}
    \label{eq:length-prior}
\end{equation}
Here $L$ denotes total link length in meters. The sampler also varies displacement directions and joint-frame rotations, assigns local-$Z$ axes and $[-2\pi,2\pi]$ limits, and exports gripper-equipped assets with planner and simulator configurations. \Cref{app:morphgen} specifies the sampling distributions and export settings.

\paragraph{Home poses and pretraining samples.}
\label{sec:morphgen-data}
A cuRobo solver~\cite{curobo} searches perturbed tabletop targets with 32 IK seeds and self-collision checking. A successful configuration is absorbed into the joint origins to define the home pose; descriptions with no solution are discarded. Unlike frame normalization, this changes the controller zero without shifting joint-limit intervals, so it need not preserve the original bounded configuration set.

Each training sample pairs a description and configuration with forward-kinematics targets from the original URDF tree, corrected into the normalized frames. Inputs contain $J\times7$ values (origin translation/rotation and joint angle), and targets contain $J\times10$ values (position, 6D rotation, and angle), with padding to seven joints. Pretraining uses 40 million morphology--state pairs.

\subsection{RoboMorpher Action Tokenization}
\label{sec:method}
Given morphology $\mathcal M$ and configuration $q\in\mathbb R^J$, RoboMorpher assigns two code indices to each valid joint. A morphology-conditioned decoder reconstructs joint angles and link poses from these codes. A mask $m$ excludes padding from encoding, quantization, and reconstruction.

\subsubsection{Morphology-Conditioned Joint Encoding}
\label{sec:encoding}
Separate two-layer Transformer branches encode the origin parameters and joint angles, retaining one feature per joint. Morphology features $c_{\mathcal M,i}$ modulate a six-layer action encoder and four-layer decoder through joint-wise AdaLN-Zero~\cite{dit}. All blocks have width 1024, eight heads, and bidirectional attention.

For head $h$, query joint $i$, and key joint $j$, the attention logit is
\begin{align}
 A^h_{ij}={Q_i^{h\top}K_j^h\over\sqrt d}
 &+b_s^h(|i-j|+1)\nonumber\\
 &+b_p^h((i-j)_+)+b_c^h((j-i)_+)+M_{ij}.
 \label{eq:attention}
\end{align}
Here $d$ is the head width and $(x)_+=\max(x,0)$. The zero-initialized tables encode hop distance and directed upstream/downstream relations; directional index zero contributes no bias. $M$ masks invalid keys, and padded features are zeroed. Thus chain biases describe joint connectivity, while AdaLN supplies morphology-dependent geometric information.

\subsubsection{Joint-Aligned Residual Quantization}
A linear map projects each conditioned feature to $u_i\in\mathbb R^{256}$. Residual vector quantization (RVQ)~\cite{rqvae} uses two 2048-entry codebooks shared across joints and descriptions. Quantization proceeds as
\begin{align}
 z_i^{(1)}&=\argmin_k\|u_i-e_k^{(1)}\|_2^2,\\
 z_i^{(2)}&=\argmin_k\|u_i-e_{z_i^{(1)}}^{(1)}-e_k^{(2)}\|_2^2,\\
 \widehat u_i&=e_{z_i^{(1)}}^{(1)}+e_{z_i^{(2)}}^{(2)}.
 \label{eq:rvq}
\end{align}
The resulting two indices per joint define a frame-level code sequence. Quantized features are projected back to width 1024 for conditional decoding. Codebooks use exponential moving averages and a straight-through gradient estimator (\cref{app:training-settings}). Decoding depends on $\mathcal M$, so a shared code need not correspond to the same joint value on every robot.

\subsubsection{Joint and Geometric Reconstruction}
\label{sec:geometry}
For each valid joint, the decoder predicts its normalized angle and child-link position, together with a 6D rotation~\cite{rotation6d}. For $N=\sum_i m_i$, the objective is
\begin{align}
 \mathcal L_q&={1\over N}\sum_i m_i|\widehat{\bar q}_i-\bar q_i|,\\
 \mathcal L_p&={1\over3N}\sum_i m_i\|\widehat{\bar p}_i-\bar p_i\|_1,\displaybreak[1]\\
 \mathcal L_R&={1\over\pi N}\sum_i m_i\,d_{\mathrm{SO}(3)}(\widehat R_i,R_i),\\
 \mathcal L_{\mathrm{tok}}&={\mathcal L_q+\mathcal L_p+0.5\mathcal L_R+\mathcal L_{\mathrm{RVQ}}}.
 \label{eq:tokloss}
\end{align}
Bars denote normalized values. $d_{\mathrm{SO}(3)}$ is the rotation geodesic in radians. $\mathcal L_{\mathrm{RVQ}}$ includes the commitment weight. Rotations are denormalized before Gram--Schmidt conversion. Per-feature min--max statistics are shared across valid joints and robots. Raw joint angles are not wrapped.

The joint and pose objectives supervise complementary aspects of each configuration: its control coordinates and its spatial realization. Because poses are predicted by an auxiliary head, this objective does not enforce exact FK consistency of decoded angles. All tokenizer parameters, including the morphology encoder, remain frozen during policy learning.

\subsection{VLA Policy Integration}
\label{sec:integration}
To use the pretrained representation for control, we combine Qwen3-VL-4B-Instruct~\cite{qwen3vl}, the frozen tokenizer, and a two-branch diffusion Transformer (DiT). Given images $I$, instruction $\ell$, state $s$, and morphology $\mathcal M$, the policy predicts 50-frame chunks of absolute joint positions and gripper commands. Each arm occupies seven padded joint slots and one gripper channel, giving 16 coordinates for two arms. Masks exclude absent joints and arms.

\subsubsection{From Frame Tokens to Action Chunks}
\label{sec:temporal}
To apply the frame-level tokenizer to action chunks, we fit $K=10$ nodes to each arm's 50-frame trajectory $A\in\mathbb R^{50\times J}$. With endpoint-aligned interpolation matrix $U\in\mathbb R^{50\times10}$, ridge fitting yields $B^*\in\mathbb R^{10\times J}$,
\begin{equation}
 B^*=(U^\top U+10^{-4}I)^{-1}U^\top A.
 \label{eq:temporalfit}
\end{equation}
Each fitted node is tokenized independently. Decoded nodes are interpolated to 50 frames at inference. Grippers use linear resampling and 256-bin quantization. An arm with $J$ joints contributes $K(2J+1)$ indices, giving at most 300 for two seven-joint arms. 

\subsubsection{Morphology-Conditioned Policy Architecture}
LayerNorm and a linear map project each VLM hidden-state level into width-1024, 16-head action blocks. The $L$ action blocks align with the VLM layers and alternate VLM cross-attention with action self-attention. Following the partially shared design of Discrete Forcing~\cite{discreteforcing}, our branches share two initial blocks and use separate remaining blocks and output heads. The branches do not exchange attention. Current state and 32 learned future queries provide context, while frozen URDF features modulate each branch through AdaLN. The VLM instruction additionally includes 256-bin state tokens.

% Queue the task showcase before the benchmark section for page 6.
\begin{figure*}[!t]
 \centering
 \includegraphics[width=\textwidth]{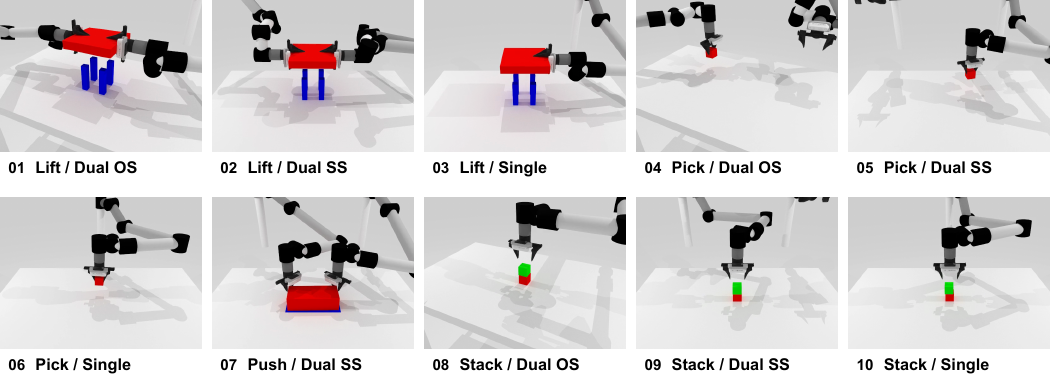}
 \caption{\textbf{The ten EmbodiBench tasks.} Successful saved-trajectory keyframes, ordered as in \cref{tab:embodibench-tasks}. SS/OS denote same-side/opposite-side dual-arm placement. These visualization replays are distinct from learned-policy rollouts.}
 \label{fig:embodibench}
\end{figure*}

\subsubsection{Discrete Prediction and Refinement}
\label{sec:policy}
The discrete branch minimizes cross-entropy over masked valid action codes, using a cosine masking ratio determined by a uniformly sampled time. At inference, eight MaskGIT iterations~\cite{maskgit} commit codes by confidence. Frozen decoding and interpolation yield the coarse trajectory $a_d$.

The continuous branch refines $a_d$ in normalized action space. For target trajectory $a^*$ and context $c=(I,\ell,s,\mathcal M)$,
\begin{align}
 x_0&=0.3a_d+0.7\epsilon,\quad \epsilon\sim\mathcal N(0,I),\\
 x_\gamma&=(1-\gamma)x_0+\gamma a^*,\\
 \mathcal L_{\mathrm{FM}}&=\mathbb E\left[\|v_\phi(x_\gamma,\gamma,c)-(a^*-x_0)\|_m^2\right].
 \label{eq:flowloss}
\end{align}
Here $\|\cdot\|_m^2$ averages squared error over valid action coordinates per example. Both trajectories use per-slot 1st/99th-percentile normalization and clipping to $[-2.2,2.2]$. Training reconstructs $a_d$ from target-trajectory codes, whereas inference uses predicted codes.

The policy minimizes $0.5\mathcal L_{\mathrm{CE}}+\mathcal L_{\mathrm{FM}}$. Four Euler steps of size $1/4$ advance $\gamma$ from zero to one, producing the 50-frame action chunk. \Cref{app:policy-settings} gives optimization and training-time sampling settings.

\section{EmbodiBench Evaluation}
\label{sec:embodibench}
EmbodiBench evaluates whether the learned representation supports manipulation across generated kinematic chains in SAPIEN. Its ten tasks comprise three lifting, three picking, three stacking, and one pushing variant, with single arms or same-/opposite-side dual arms (\cref{fig:embodibench}).

\paragraph{Data and coverage.}
Each task uses 30 URDFs and 50 demonstrations per task--URDF pair, totaling 15,000 demonstrations. The overlapping task-specific sets contain 55 distinct generated morphologies. Dual-arm trials use the same morphology for both arms. Policies receive head and wrist RGB views resized to $224\times224$. Absent arm channels are masked.

\paragraph{Evaluation.}
Evaluation uses separate scene seeds and five trials per task--URDF pair: 150 per task and 1,500 per method. Scenes rejected by the expert planner do not consume the quota. Success rates are therefore conditional on planner feasibility. Each accepted trial has a fixed instruction and permits 300 executed action commands, counted per frame within a chunk. We report task success rates and their unweighted mean.

\paragraph{Success criteria.}
Picking requires an object height increase exceeding 7\,cm. Pushing requires planar placement within 3\,cm per coordinate and orientation error below $9^\circ$. Stacking requires alignment within $(2.5,2.5,1.2)$\,cm of the target pose and open grippers. Lifting requires height above 0.9\,m, a rise of at least 3\,cm, and a valid two-finger grasp by at least one arm for 15 consecutive physics steps (\cref{app:benchmark-details}).

\section{Experiments}
\label{sec:experiments}
We evaluate downstream task performance, the contribution of representation components, and reconstruction fidelity on URDFs unseen during tokenizer pretraining.

\subsection{Experimental Setup}
\label{sec:experimental-setup}
\paragraph{Training and evaluation.}
The tokenizer is pretrained on 40 million morphology--state pairs for 300 epochs using eight GPUs and a global batch of 32,768. All downstream URDFs are excluded. Policy training updates the VLM and action head with global batch 128 while keeping the tokenizer frozen. RoboTwin~2.0 uses 3,750 demonstrations ($15$ tasks, $5$ embodiments, $50$ demonstrations each) for 30k steps. EmbodiBench uses 15,000 for 60k steps. Full settings appear in \cref{app:training-settings}. RoboTwin~2.0 evaluates 20 trials per task--embodiment pair. EmbodiBench follows \cref{sec:embodibench}. Both benchmarks use 1,500 trials per method with equal task quotas, so macro-averaged and pooled success agree up to rounding.

\paragraph{Comparisons.}
\emph{StarVLA (joint)} predicts continuous joint actions. \emph{256-bin} uses uniform bins, MaskGIT, and ten-node fitting. \emph{FAST}~\cite{fast} uses autoregressive codes without fitting. \emph{Angle+M} removes the link-pose prediction head and its reconstruction loss. \emph{Angle} also removes morphology conditioning and graph biases and changes commitment weight from 0.1 to 0.25. \emph{StarVLA (EE)} is a separate control-interface reference. Ours denotes the policy using the full RoboMorpher tokenizer.

% Generated by scripts/build_tables.py; edit the source/script.
\begin{table*}[t]
\centering
\small
\setlength{\tabcolsep}{4pt}
\caption{\textbf{Full RoboTwin~2.0 results. Success rate (\%).} All 15 evaluated tasks across five embodiments. 20 trials per pair. Bold/underline mark best/second-best positive joint-space scores (ties included). EE uses a separate control interface.}
\label{tab:robotwin-tasks}
\normalsize
\renewcommand{\arraystretch}{1.05}
\begin{tabular*}{\textwidth}{@{\extracolsep{\fill}}lrrrrrrr@{}}
\toprule
& \multicolumn{3}{c}{Joint-space baselines} & \multicolumn{2}{c}{Representation variants} & Full model & Reference \\ \cmidrule(lr){2-4}\cmidrule(lr){5-6}\cmidrule(lr){7-7}\cmidrule(l){8-8}Task & Joint & 256-bin & FAST & Angle & Angle+M & Ours & EE \\
\midrule
grab\_roller & 90.00\% & \textbf{97.00\%} & \underline{95.00\%} & \underline{95.00\%} & \textbf{97.00\%} & 92.00\% & 100.00\% \\
shake\_bottle & 92.00\% & \underline{95.00\%} & \textbf{96.00\%} & 88.00\% & \textbf{96.00\%} & 94.00\% & 100.00\% \\
handover\_mic & 47.00\% & 67.00\% & \underline{74.00\%} & 52.00\% & \underline{74.00\%} & \textbf{75.00\%} & 93.00\% \\
place\_burger\_fries & 59.00\% & 65.00\% & \underline{67.00\%} & 58.00\% & 53.00\% & \textbf{74.00\%} & 52.00\% \\
place\_empty\_cup & \underline{69.00\%} & \textbf{70.00\%} & 68.00\% & \underline{69.00\%} & \underline{69.00\%} & \textbf{70.00\%} & 85.00\% \\
click\_bell & \underline{88.00\%} & 72.00\% & 69.00\% & 84.00\% & \textbf{100.00\%} & 84.00\% & 88.00\% \\
lift\_pot & 28.00\% & \textbf{37.00\%} & 17.00\% & 5.00\% & 22.00\% & \underline{30.00\%} & 82.00\% \\
open\_laptop & 66.00\% & 73.00\% & \underline{76.00\%} & 64.00\% & 70.00\% & \textbf{82.00\%} & 97.00\% \\
beat\_block\_hammer & 10.00\% & 13.00\% & \underline{25.00\%} & 11.00\% & 18.00\% & \textbf{34.00\%} & 32.00\% \\
stack\_bowls\_three & 25.00\% & \underline{59.00\%} & 54.00\% & 54.00\% & \textbf{64.00\%} & 54.00\% & 68.00\% \\
pick\_dual\_bottles & \underline{32.00\%} & \textbf{35.00\%} & \underline{32.00\%} & 25.00\% & 30.00\% & 29.00\% & 65.00\% \\
move\_can\_pot & 20.00\% & \underline{21.00\%} & 17.00\% & 20.00\% & 12.00\% & \textbf{26.00\%} & 52.00\% \\
place\_phone\_stand & 31.00\% & 34.00\% & \underline{42.00\%} & 27.00\% & 39.00\% & \textbf{50.00\%} & 51.00\% \\
put\_bottles\_dustbin & \textbf{21.00\%} & 10.00\% & 12.00\% & 8.00\% & \underline{18.00\%} & 17.00\% & 42.00\% \\
place\_dual\_shoes & 22.00\% & \underline{26.00\%} & 17.00\% & 20.00\% & 16.00\% & \textbf{35.00\%} & 42.00\% \\
\midrule
\textbf{Mean} & 46.67\% & 51.60\% & 50.73\% & 45.33\% & \underline{51.87\%} & \textbf{56.40\%} & 69.93\% \\
\bottomrule
\end{tabular*}
\end{table*}

\subsection{Multi-Embodiment Task Performance}
\label{sec:policy-results}
RoboMorpher achieves the highest mean success among the evaluated joint-space methods: 56.40\% on RoboTwin~2.0 and 23.27\% on EmbodiBench (\cref{tab:robotwin-tasks,tab:embodibench-tasks}), exceeding StarVLA (joint) by 9.73 and 15.34 percentage points (pp).

\paragraph{Gains across RoboTwin~2.0 tasks.}
RoboMorpher improves over continuous joint control on 12 of 15 tasks. The largest gains are on bowl stacking (54\% vs.\ 25\%), microphone handover (75\% vs.\ 47\%), and hammering (34\% vs.\ 10\%). Success decreases by 3--4\,pp on bell clicking, dual-bottle picking, and bottle disposal. Against 256-bin and FAST, the mean gains are 4.80 and 5.67\,pp, with improvements on nine and eleven tasks, respectively. 

% Generated by scripts/build_tables.py; edit the source/script.
\begin{table*}[t]
\centering
\small
\setlength{\tabcolsep}{4pt}
\caption{\textbf{Full EmbodiBench results. Success rate (\%).} All 10 tasks with 30 URDFs per task. Five trials per pair. Bold/underline mark best/second-best positive joint-space scores (ties included). EE uses a separate control interface.}
\label{tab:embodibench-tasks}
\normalsize
\renewcommand{\arraystretch}{1.05}
\begin{tabular*}{0.9\textwidth}{@{\extracolsep{\fill}}lrrrrrrr@{}}
\toprule
& \multicolumn{3}{c}{Joint-space baselines} & \multicolumn{2}{c}{Representation variants} & Full model & Reference \\ \cmidrule(lr){2-4}\cmidrule(lr){5-6}\cmidrule(lr){7-7}\cmidrule(l){8-8}Task & Joint & 256-bin & FAST & Angle & Angle+M & Ours & EE \\
\midrule
lift\_dual\_os & \underline{14.67\%} & 14.00\% & 9.33\% & 10.67\% & 10.67\% & \textbf{61.33\%} & 62.67\% \\
lift\_dual\_ss & \underline{14.00\%} & 4.00\% & 11.33\% & 6.67\% & 8.67\% & \textbf{52.67\%} & 58.67\% \\
lift\_single & \underline{13.33\%} & 11.33\% & 10.67\% & 6.00\% & 9.33\% & \textbf{51.33\%} & 78.67\% \\
\addlinespace[2pt]
pick\_dual\_os & \underline{6.00\%} & 2.67\% & 4.00\% & 1.33\% & 3.33\% & \textbf{23.33\%} & 14.67\% \\
pick\_dual\_ss & \underline{6.00\%} & 4.00\% & 2.67\% & 2.67\% & 4.67\% & \textbf{11.33\%} & 8.00\% \\
pick\_single & \textbf{9.33\%} & 4.00\% & \underline{5.33\%} & \underline{5.33\%} & 1.33\% & 4.00\% & 32.67\% \\
\addlinespace[2pt]
push\_dual\_ss & \underline{16.00\%} & 15.33\% & 15.33\% & 13.33\% & 14.67\% & \textbf{28.67\%} & 62.67\% \\
\addlinespace[2pt]
stack\_dual\_os & 0.00\% & 0.00\% & 0.00\% & 0.00\% & 0.00\% & 0.00\% & 1.33\% \\
stack\_dual\_ss & 0.00\% & 0.00\% & 0.00\% & 0.00\% & \textbf{0.67\%} & 0.00\% & 0.00\% \\
stack\_single & 0.00\% & 0.00\% & \textbf{0.67\%} & 0.00\% & 0.00\% & 0.00\% & 0.67\% \\
\midrule
\textbf{Mean} & \underline{7.93\%} & 5.53\% & 5.93\% & 4.60\% & 5.33\% & \textbf{23.27\%} & 32.00\% \\
\bottomrule
\end{tabular*}
\end{table*}

% Generated by scripts/build_tables.py; edit the source/script.
\begin{table*}[t]
\centering
\small
\setlength{\tabcolsep}{4pt}
\caption{\textbf{Tokenizer reconstruction.} All saved task frames across six description groups. Geometry is predicted directly by the decoder. P90 is the pooled 90th percentile. Bold/underline mark the lowest/second-lowest value per column (ties included). Metric definitions appear in \cref{app:training-settings}.}
\label{tab:reconstruction}
\normalsize
\renewcommand{\arraystretch}{1.05}
\begin{tabular*}{\textwidth}{@{\extracolsep{\fill}}lrrrrrrrrr@{}}
\toprule
& \multicolumn{3}{c}{Joint angle ($^\circ$)} & \multicolumn{3}{c}{Position (m)} & \multicolumn{3}{c}{Orientation ($^\circ$)} \\ \cmidrule(lr){2-4}\cmidrule(lr){5-7}\cmidrule(l){8-10} Description & Mean & P90 & Max & Mean & P90 & Max & Mean & P90 & Max \\
\midrule
ALOHA & \underline{0.4124} & 0.8951 & \underline{14.4648} & \underline{0.004255} & \underline{0.009446} & 0.116570 & \underline{1.3237} & \underline{3.4343} & 88.6863 \\
ARX-X5 & \textbf{0.4107} & \textbf{0.7768} & 52.6336 & 0.004899 & 0.013960 & \underline{0.112769} & \textbf{1.0860} & \textbf{2.0659} & 133.9178 \\
Franka Panda & 0.4563 & 1.0207 & 42.7852 & 0.009204 & 0.022782 & 0.644741 & 1.8655 & 4.4817 & 178.8929 \\
Piper & 0.4448 & \underline{0.8695} & 40.3280 & \textbf{0.003503} & \textbf{0.007992} & 0.186133 & 1.6299 & 4.5327 & \underline{88.5053} \\
UR5-WSG & 0.5429 & 1.2653 & \textbf{5.6463} & 0.007017 & 0.016207 & \textbf{0.048518} & 1.3591 & 3.7680 & \textbf{11.4713} \\
Generated URDFs & 0.4764 & 1.0696 & 80.9388 & 0.008355 & 0.018587 & 0.752892 & 2.1928 & 4.6185 & 179.9428 \\
\bottomrule
\end{tabular*}
\end{table*}

\paragraph{Lifting gains on EmbodiBench.}
The three lifting tasks average 55.11\% success, compared with 14.00\% for StarVLA (joint), accounting for about 80\% of the overall gain. Picking improves from 7.11\% to 12.89\% on average, and pushing from 16.00\% to 28.67\%. The policy improves on six of seven non-stacking tasks, but single-arm picking falls from 9.33\% to 4.00\%. All three stacking rates remain zero, and no evaluated method exceeds 1.33\% on these tasks. This contrasts with 54\% success on RoboTwin~2.0 bowl stacking, but differences in objects and success predicates preclude attributing the gap solely to morphology.

Opposite-side trials yield higher lifting (61.33\% vs.\ 52.67\%) and picking (23.33\% vs.\ 11.33\%) success. Because placement and scene geometry vary together, this does not isolate an effect of arm placement.

\paragraph{Tokenization baselines.}
256-bin and FAST exceed continuous joint control on RoboTwin~2.0 but trail it on EmbodiBench. Their EmbodiBench lifting rates are 9.78\% and 10.44\%, compared with 55.11\% for RoboMorpher. Temporal processing and decoding also differ between these baselines; the link-pose ablation below offers a more controlled comparison.

\paragraph{Comparison with EE control.}
The EE reference achieves 69.93\% and 32.00\% success, exceeding RoboMorpher by 13.53 and 8.73\,pp. On EmbodiBench, RoboMorpher performs better on both dual-arm picking tasks but trails by 28.67\,pp on single-arm picking and 34.00\,pp on pushing. This gap includes differences in execution interface and cannot be attributed solely to representation quality.

\subsection{Effect of Representation Design}
\label{sec:ablation-results}
\paragraph{Link-pose supervision.}
Removing link-pose reconstruction reduces mean success by 4.53\,pp on RoboTwin~2.0 and 17.94\,pp on EmbodiBench. On EmbodiBench, lifting drops from 55.11\% to 9.56\% on average, and all seven non-stacking tasks decline. Angle+M retains morphology input yet reaches only 5.33\% overall, below continuous joint control at 7.93\%. These results support the value of link-pose supervision on generated morphologies, particularly for lifting.

Link-pose supervision improves nine RoboTwin~2.0 tasks, including burger-and-fries (+21\,pp) and shoe placement (+19\,pp), but reduces bell clicking and bowl stacking by 16 and 10\,pp. Its benefit is therefore task-dependent. 

\paragraph{Morphology conditioning and graph biases.}
Angle+M exceeds Angle by 6.54\,pp on RoboTwin~2.0 and 0.73\,pp on EmbodiBench. Morphology conditioning, graph biases, and commitment weight change together, preventing attribution to individual components. Continuous refinement is not separately ablated.

\subsection{Reconstruction Fidelity}
\label{sec:recon-results}
Reconstruction covers every frame of 3,750 RoboTwin~2.0 and 15,000 EmbodiBench policy-training demonstrations, using URDFs unseen during tokenizer pretraining (\cref{app:reconstruction-data}). We pool valid joint/link errors by description, comparing decoded angles with inputs and decoded poses with input-angle FK targets.

Mean joint errors span $0.4107^\circ$--$0.5429^\circ$ (\cref{tab:reconstruction}). Generated URDFs average $0.4764^\circ$ joint error but have the highest mean orientation error ($2.1928^\circ$). Their maximum position and orientation errors reach 0.752892\,m and $179.9428^\circ$, revealing substantial outliers. These sample-count-dependent maxima do not establish relative tail robustness. Raw-angle errors retain periodic-boundary differences.

\begin{figure}[h!]
    \centering
    \includegraphics[width=\columnwidth]{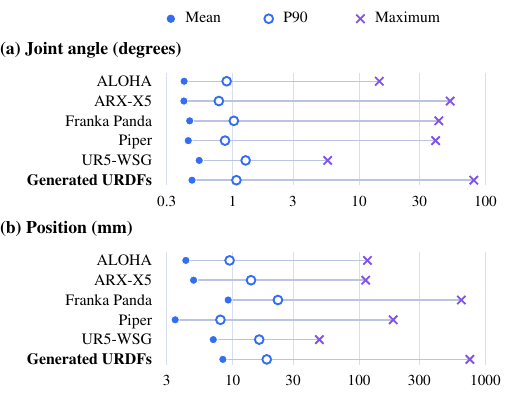}
    \caption{\textbf{Reconstruction error profiles.} Mean, P90, and maximum joint-angle and position errors from \cref{tab:reconstruction}. Both axes are logarithmic.}
    \label{fig:reconstruction-profile}
\end{figure}

Reconstruction weights joint/link samples equally; policy success weights tasks equally. Representation fidelity alone establishes neither action-selection quality nor FK consistency of decoded angles.

\subsection{Limitations}
Unseen-URDF reconstruction does not establish zero-shot policy transfer: downstream policy training and evaluation lack a disjoint-morphology split. Survey coverage is limited, stacking remains difficult, and reconstruction has large outliers. Scaling behavior and variability across training seeds remain unmeasured (\cref{app:tasks}).

\section{Conclusion}
RoboMorpher learns morphology-conditioned action codes from MorphGen kinematic supervision and improves mean joint-space policy success on RoboTwin~2.0 and EmbodiBench. These findings support kinematic pretraining followed by task-specific policy learning. Ablations support link-pose supervision, particularly for lifting, while reconstruction exposes substantial outliers despite low average errors on unseen tokenizer URDFs. Improving stacking, reducing reconstruction tails, and testing policy transfer under disjoint morphology splits are important directions for future work.

% References begin after the eight-page main text.
\clearpage
{
    \small
    \bibliographystyle{ieeenat_fullname}
    \bibliography{main}
}
% Keep the appendix in the public preprint to resolve all cross-references.
\clearpage
\appendix
\raggedbottom
\maketitlesupplementary

\section{Robot-Description Survey}
\label{app:survey}
% Generated by scripts/build_tables.py; edit the source/script.
\begin{table*}[t]
\centering
\small
\setlength{\tabcolsep}{4pt}
\caption{\textbf{Complete description inventory of structure and scale.} All entries are serial manipulators. DoF is the reported arm DoF, and masses and reaches are approximate description-level values. Joint types are R (revolute), C (continuous), P (prismatic), F (fixed), H (hinge), and S (slide/slider). Limits are B (bounded in the stated scope), M (mixed), and U (unrestricted). A dash denotes an unavailable value.}
\label{tab:survey-inventory}
\begin{tabular}{@{}rlcrlcrr@{}}
\toprule
ID & Name & Format & DoF & Joint types & Limits & Mass (kg) & Reach (m) \\
\midrule
1 & e.DO & URDF & 6 & All C & U & 0.5 & -- \\
2 & FR3 & MJCF & 7 & All H & B & 15.49 & 0.85 \\
3 & Gen2 & URDF & 6 & C:3; R:3; F:6 & M & 4.35 & 0.96 \\
4 & Gen3 Lite & URDF & 6 & R:10; F:4 & B & 5.56 & 0.79 \\
5 & Gen3 & MJCF & 7 & All H & M & 6.34 & 0.93 \\
6 & Gen3 & URDF & 6 & C:3; R:3; F:4 & M & 6.14 & 0.93 \\
7 & iiwa 14 & MJCF & 7 & All H & B & 29.26 & 0.82 \\
8 & iiwa 7 & URDF & 7 & R:7; F:1 & B & 26.96 & 0.80 \\
9 & Koch v1.1 & MJCF & 6 & H:5; S:1 & B & 0.226 & 0.25 \\
10 & Koch v1.1 & URDF & 6 & All R & B & 0.29 & 0.25 \\
11 & Lite 6 & URDF & 6 & R:6; F:1 & B & 6.22 & 0.60 \\
12 & Low-Cost Robot Arm & URDF & 5 & All C & U & 2.23 & 0.25 \\
13 & M-710iC & URDF & 6 & R:6; F:3 & B & -- & 2.05 \\
14 & Med 14 & URDF & 7 & R:7; F:1 & B & 31.65 & 1.30 \\
15 & Med 7 & URDF & 7 & R:7; F:1 & B & 27.18 & 0.70 \\
16 & Panda & MJCF & 7 & H:7; S:2 & B & 18.17 & 0.85 \\
17 & Panda & URDF & 7 & R:7; P:2; F:3 & B & 18.15 & 0.85 \\
18 & PiPER & URDF & 6 & R:6; P:2; F:2 & B & 4.24 & 0.60 \\
19 & Poppy Ergo Jr & URDF & 6 & C:6; F:1 & U & 0.6 & 0.30 \\
20 & Sawyer & MJCF & 6 & All H & B & 24.21 & 1.0 \\
21 & SO-ARM 100 & URDF & 5 & All R & B & 1.61 & 0.35 \\
22 & UR10 & URDF & 6 & R:6; F:4 & B & 32.72 & 1.30 \\
23 & UR10e & MJCF & 6 & All H & B & 33.5 & 1.30 \\
24 & UR3 & URDF & 6 & R:6; F:4 & B & 11.63 & 0.50 \\
25 & UR3e & URDF & 6 & R:6; F:3 & B & 11.23 & 0.50 \\
26 & UR5 & URDF & 6 & R:6; F:4 & B & 22.77 & 0.85 \\
27 & UR5e & MJCF & 6 & All H & B & 22.82 & 0.85 \\
28 & ViperX 300 & URDF & 6 & R:6; P:2; F:5; C:1 & M & 3.47 & 0.75 \\
29 & WidowX 250 & MJCF & 6 & H:6; S:2 & B & 2.88 & 0.60 \\
30 & xArm5 & URDF & 5 & R:5; F:1 & B & 8.54 & 0.70 \\
31 & xArm6 & URDF & 6 & R:6; F:1 & B & 9.53 & 0.70 \\
32 & xArm7 & URDF & 7 & R:7; F:1 & B & 5.50 & 0.70 \\
33 & Z1 & URDF & 6 & R:6; F:1 & B & 4.43 & 0.80 \\
\bottomrule
\end{tabular}
\end{table*}

% Generated by scripts/build_tables.py; edit the source/script.
\begin{table*}[t]
\centering
\small
\setlength{\tabcolsep}{4pt}
\caption{\textbf{Recorded tools and sensing.} IDs match \cref{tab:survey-inventory}. The 19 omitted entries list neither a tool nor sensing. Tool DoF and couplings follow the survey and need not equal independent actuation. Missing records concern the description, not the physical robot. The AR tag is a passive marker.}
\label{tab:survey-tools}
\begin{tabular}{@{}rlclll@{}}
\toprule
ID & Name & Format & Tool & Reported tool DoF & Sensor / marker \\
\midrule
3 & Gen2 & URDF & Two-finger gripper & 0 & None listed \\
4 & Gen3 Lite & URDF & Two-finger, four-joint gripper & 1 & None listed \\
5 & Gen3 & MJCF & None listed & -- & Force/torque at EE \\
6 & Gen3 & URDF & None listed & -- & Camera at EE \\
9 & Koch v1.1 & MJCF & Single-sided gripper & 1 translational & None listed \\
10 & Koch v1.1 & URDF & Single-sided gripper & 1 rotational & None listed \\
12 & Low-Cost Robot Arm & URDF & Single-sided gripper & 1 continuous rotation & None listed \\
16 & Panda & MJCF & Two-finger translating gripper & 2 translational & None listed \\
17 & Panda & URDF & Two-finger translating gripper & 1 translational; mimic & None listed \\
18 & PiPER & URDF & Two-finger translating gripper & 2 translational & None listed \\
19 & Poppy Ergo Jr & URDF & Lamp & -- & None listed \\
21 & SO-ARM 100 & URDF & Single-sided gripper & 1 rotational & None listed \\
28 & ViperX 300 & URDF & Two-finger translating gripper & 1 translational; mimic & AR tag at EE \\
29 & WidowX 250 & MJCF & Two-finger translating gripper & 1 translational; opposing fingers & None listed \\
\bottomrule
\end{tabular}
\end{table*}

% Generated by scripts/build_tables.py; edit the source/script.
\begin{table*}[t]
\centering
\small
\setlength{\tabcolsep}{4pt}
\caption{\textbf{Link-length records as supplied (m).} Blank source cells are shown as dashes. Reported zeros are retained. These records are kept separate from the main inventory because description identities and extraction conventions are not fully aligned. They are not treated as validated workspace radii or summed to estimate reach.}
\label{tab:survey-links}
\begin{tabular}{@{}lrrrrrrr@{}}
\toprule
Record & 1 & 2 & 3 & 4 & 5 & 6 & 7 \\
\midrule
eDO & 0.145292 & 0.18967 & 0.210498 & 0.159044 & 0.1053 & 0.163646 & -- \\
Gen2 & 0.15675 & 0.11876 & 0.410 & 0.2076 & 0.10375 & 0.10375 & -- \\
Gen3 Lite & 0.12825 & 0.11885 & 0.280 & 0.14142 & 0.1088 & 0.1088 & -- \\
Gen3 & 0.15643 & 0.12849 & 0.410 & 0.20853 & 0.10593 & 0.10593 & -- \\
iiwa14 & 0.1575 & 0.2025 & 0.2045 & 0.2155 & 0.1845 & 0.2155 & 0.0810 \\
iiwa7 & 0.15 & 0.19 & 0.21 & 0.19 & 0.2 & 0.19946 & 0.10122 \\
koch-v1-1 & 0.04262 & 0.02596 & 0.10950 & 0.10052 & 0.04687 & 0.02096 & -- \\
M-710iC & 0.565 & 0.150 & 0.870 & 0.170 & 1.016 & 0.175 & -- \\
PIPER & 0.123 & 0.0 & 0.28503 & 0.25171 & 0.0 & 0.091 & -- \\
SO-ARM 100 & 0.04812 & 0.10697 & 0.11600 & 0.13500 & 0.06010 & -- & -- \\
UR10 & 0.1273 & 0.220941 & 0.63568 & 0.5723 & 0.1149 & 0.1157 & -- \\
UR3 & 0.1519 & 0.1198 & 0.26062 & 0.21325 & 0.08505 & 0.08535 & -- \\
UR5 & 0.089159 & 0.13585 & 0.44153 & 0.3922 & 0.093 & 0.09465 & -- \\
Z1 & 0.0585 & 0.045 & 0.35 & 0.22533 & 0.07 & 0.0492 & -- \\
panda & 0.333 & 0.0 & 0.316 & 0.0825 & 0.39276 & 0.0 & 0.088 \\
Gen3 (MJCF) & 0.15643 & 0.12849 & 0.21048 & 0.21048 & 0.20853 & 0.10593 & 0.10593 \\
Lite 6 (MJCF) & 0.2435 & 0.0 & 0.2002 & 0.24367 & 0.0 & 0.0625 & -- \\
Poppy Ergo Jr & 0.0328 & 0.0240 & 0.0540 & 0.0450 & 0.0480 & 0.0540 & -- \\
Sawyer (MJCF) & 0.08 & 0.25540 & 0.19977 & 0.26337 & 0.17784 & 0.27674 & 0.15228 \\
Low-Cost Robot Arm & 0.03900 & 0.01730 & 0.10935 & 0.09051 & 0.04502 & -- & -- \\
Med 14 & 0.14700 & 0.21273 & 0.22822 & 0.19229 & 0.20777 & 0.20507 & 0.11524 \\
Med 7 & 0.14750 & 0.19279 & 0.20777 & 0.19279 & 0.20777 & 0.20507 & 0.11524 \\
UR10e (MJCF) & 0.18100 & 0.17600 & 0.62812 & 0.57100 & 0.13500 & 0.12000 & -- \\
UR3e & 0.15200 & 0.12000 & 0.26112 & 0.21300 & 0.10400 & 0.08500 & -- \\
UR5e (MJCF) & 0.16300 & 0.13800 & 0.44473 & 0.39200 & 0.12700 & 0.10000 & -- \\
ViperX 300 (MJCF) & 0.07900 & 0.04805 & 0.30585 & 0.20000 & 0.10000 & 0.06974 & -- \\
WidowX 250 (MJCF) & 0.07200 & 0.03865 & 0.25490 & 0.17500 & 0.07500 & 0.06500 & -- \\
xArm7 & 0.26700 & 0.0 & 0.29300 & 0.05250 & 0.35116 & 0.0 & 0.12323 \\
xArm6 & 0.26700 & 0.0 & 0.28949 & 0.35116 & 0.0 & 0.12323 & -- \\
xArm5 & 0.26700 & 0.0 & 0.28949 & 0.35116 & 0.12323 & -- & -- \\
\bottomrule
\end{tabular}
\end{table*}

\subsection{Inventory and Scope}
\Cref{tab:survey-inventory} lists all 33 descriptions of serial manipulators, covering 30 names. Gen3, Koch v1.1, and Panda each have two entries. Distinct model names, such as UR5 and UR5e, remain separate. Mass, reach, tool DoF, and sensing are transcribed description properties, not independently validated hardware measurements.

Arm DoF follows the explicit survey field. Joint-type counts may include fixed and tool joints and therefore cannot replace that field. Likewise, mimic or other coupling constraints can make the number of movable fingers differ from the number of independent control dimensions. Chain extraction and action masks use independent arm joints.

\subsection{Limit and Tool Conventions}
The B/M/U labels retain each entry's stated scope, which may cover the arm or all joints. There are 26 bounded, four mixed, and three unrestricted entries. Gen2 mixes unrestricted base/wrist with bounded shoulder/elbow joints. Gen3 MJCF lists unrestricted base, elbow-1, wrist-1, and wrist-3 joints. Its URDF entry omits elbow-1 from this list. ViperX 300 lists an unrestricted gripper joint alongside bounded arm/finger joints. The unrestricted entries are e.DO, Low-Cost Robot Arm, and Poppy Ergo Jr.

\Cref{tab:survey-tools} retains the 14 entries with a recorded tool, sensor, or marker. The other 19 list neither. Eleven entries have grippers, comprising five translating two-finger, four single-sided, one two-finger, and one two-finger four-joint. Poppy Ergo Jr lists a lamp. Sensing is recorded only for Gen3 MJCF (force/torque) and Gen3 URDF (camera). ViperX 300 lists a passive AR marker. Missing records do not imply absent hardware or describe the observation setup used in policy experiments.

\subsection{Link Records and Data Ambiguities}
\Cref{tab:survey-links} preserves all 30 link records, including reported zeros and missing entries. This table is not in one-to-one correspondence with the description inventory. FR3 has no named link record, Gen3 has unsuffixed and MJCF records, and other repeated names may have only one link record. Without the source transforms and extraction rule, these values cannot be treated as physical measurements or summed to estimate reach.

We retain five unresolved discrepancies. Gen3 reports six arm DoF in URDF and seven in MJCF. Koch v1.1 reports six arm DoF but five hinges and one slider in MJCF. Sawyer reports six arm DoF but seven link values. Lite 6 and ViperX 300 link records are labeled MJCF, whereas their inventory entries are URDF. Panda reports different tool DoF across formats, with explicit mimic coupling in URDF. The survey alone cannot distinguish hardware variants from modeling or counting differences.

\section{MorphGen Implementation}
\label{app:morphgen}
\subsection{Survey-Derived Sampling Parameters}
% Generated by scripts/build_tables.py; edit the source/script.
\begin{table}[t]
\centering
\small
\caption{\textbf{Survey-derived link prior.} Means and sample standard deviations reproduce the generator parameters to four decimals. Statistics use the supplied link records, including zeros and excluding blanks. Sampling and global rescaling change the final length distribution.}
\label{tab:morphgen-prior}
\begin{tabular}{@{}rrrr@{}}
\toprule
Joint position & Records & $\mu_i$ (m) & $\sigma_i$ (m) \\
\midrule
1 & 30 & 0.1576 & 0.1071 \\
2 & 30 & 0.1035 & 0.0805 \\
3 & 30 & 0.3029 & 0.1715 \\
4 & 30 & 0.2270 & 0.1282 \\
5 & 30 & 0.1532 & 0.1862 \\
6 & 27 & 0.1112 & 0.0677 \\
7 & 8 & 0.1103 & 0.0222 \\
\bottomrule
\end{tabular}
\end{table}

\Cref{tab:morphgen-prior} gives the seven Gaussian parameters computed from \cref{tab:survey-links}. Each column uses its non-missing values, including zeros, to estimate the mean and sample standard deviation. These are linkwise sampling priors, not workspace statistics.

For each arm, the generator samples an integer joint count uniformly from the configured inclusive interval, whose default is $[6,7]$. Each provisional length is drawn from $\mathcal N(\mu_i,\sigma_i^2)$ truncated below at zero and above at $\max(u,\mu_i+0.1)$ meters. The nominal upper bound $u$ is 0.65 with probability 0.95 and 0.35 otherwise. The sampled lengths are rescaled jointly using \cref{eq:length-prior}. Consequently, the final lengths need not retain the pre-rescaling truncation bounds.

The first displacement is along positive $Z$. Later displacements select one, two, or three active coordinates with probabilities $(0.60,0.39,0.01)$. A single active coordinate is selected from $(x,y,z)$ with probabilities $(0.15,0.15,0.70)$. A coordinate pair is selected from $(xz,yz,xy)$ with probabilities $(0.45,0.45,0.10)$. Active components are sampled from $\mathcal U(0,1)$ and normalized to unit length. The implementation rejects a direction within $5^\circ$ of the preceding direction's antiparallel orientation in the common frame. Roll and pitch initially take values in $\{0,\pm\pi/2,\pi\}$, and yaw is zero. Internal roll and pitch each independently receive a Gaussian perturbation of standard deviation 0.09 radians with probability 0.05. The terminal orientation filter accepts a downward or horizontal terminal local-$Z$ direction, expressed by the updated rotation's $R_{zz}\leq0$ condition.

The exporter assigns arm limits $[-2\pi,2\pi]$, effort 100, and velocity 5, attaches an ALOHA gripper, and writes planner and simulator configurations. Collision spheres have radius 0.035\,m and spacing 0.02\,m. The 96 examples in \cref{fig:morphgen-gallery}(b) illustrate variation, without identifying evaluation URDFs or establishing hardware feasibility.

\subsection{Inverse-Kinematics Home-Pose Selection}
The home-pose stage constructs a tabletop target grid with $x\in[0.20,0.38]$, $y\in[-0.25,0.25]$, and $z\in[0.18,0.51]$ meters, at spacings 0.03, 0.05, and 0.03\,m, respectively. Positions receive independent Gaussian noise with standard deviation 0.01\,m. Targets are shuffled and tested in batches of 50. cuRobo uses 32 IK seeds and self-collision checking. The first batch containing a successful target provides a randomly selected successful solution. A description is rejected if all candidates fail.

Each joint origin rotation absorbs $R_{a_i}(q_i^{\mathrm{home}})$ to set the controller zero. Unlike the axis-frame change in \cref{eq:localz}, this changes the numerical configuration coordinates. Because limits are not shifted, the original bounded configuration set need not be preserved. The IK filter checks sampled targets under the collision model. It does not guarantee collision-free motion throughout the joint space.

\subsection{Feature Extraction and Frames}
The feature extractor receives an explicit joint order. It composes fixed paths between consecutive selected movable joints, rejects skipped movable joints and selected mimic/prismatic joints, and expresses each selected rotational axis as local $Z$. The default representation is relative to the arm root. An optional setting can incorporate the root-to-arm home transform into the first origin. Each valid joint supplies translation and roll--pitch--yaw, plus its raw joint angle. A Boolean mask separates real joints from padding up to seven positions.

Forward kinematics on the original graph provides targets, subsequently corrected into the normalized link frames. These targets comprise position, two rotation-matrix columns, and angle. Gripper actions are handled separately. Local-$Z$ normalization removes explicit axis variation and intervening fixed paths, but chain selection, coordinate gauges, and roll--pitch--yaw conventions prevent a unique representation of all physically equivalent descriptions.

\section{Tokenizer Training and Reconstruction}
\label{app:training-settings}
% Generated by scripts/build_tables.py; edit the source/script.
\begin{table*}[t]
\centering
\small
\setlength{\tabcolsep}{4pt}
\caption{\textbf{Training data and optimization.} Tokenizer data count morphology--state pairs. Policy data count demonstrations. Durations and global batches describe completed runs. The remaining optimizer settings are from the reference configurations. Reconstruction corpus sizes are reported separately in \cref{tab:reconstruction-coverage}.}
\label{tab:optimization}
\begin{tabular}{@{}llll@{}}
\toprule
Setting & Tokenizer & VLA: RoboTwin~2.0 & VLA: EmbodiBench \\
\midrule
Training data & 40M pairs & 3,750 demonstrations & 15,000 demonstrations \\
Duration & 300 epochs & 30,000 steps & 60,000 steps \\
Global batch & 32,768 & 128 & 128 \\
Gradient accumulation & 1 & 1 & 1 \\
\midrule
Optimizer & AdamW & AdamW & AdamW \\
Learning rate & $10^{-5}$ & VLM $10^{-5}$; head $10^{-4}$ & VLM $10^{-5}$; head $10^{-4}$ \\
Adam betas & $(0.9,0.999)$ & $(0.9,0.95)$ & $(0.9,0.95)$ \\
Weight decay & $10^{-2}$ & $10^{-8}$ & $10^{-8}$ \\
Warmup & 1000 steps & 5000 steps & 5000 steps \\
Schedule / minimum LR & Cosine / $10^{-6}$ & Cosine / $5\times10^{-7}$ & Cosine / $5\times10^{-7}$ \\
Gradient norm clip & 1 & 1 & 1 \\
Precision / seed & BF16 / 42 & BF16 / 42 & BF16 / 42 \\
Validation fraction & 0.001 & 0 & 0 \\
\bottomrule
\end{tabular}
\end{table*}

\subsection{Network, Quantization, and Normalization}
The morphology and angle branches each have two Transformer layers, followed by six conditional encoder and four decoder layers (width 1024, eight heads). RVQ uses dimension 256 and two separate 2048-entry codebooks shared across joints. It uses EMA decay 0.8, 100 $k$-means initialization iterations, commitment weight 0.1, dead-code fraction 0.1, and straight-through gradients. Rotation-trick gradients, entropy regularization, and autoencoder-only warmup are disabled. Conditional stacks use joint-wise AdaLN-Zero. Spatial and directional chain-bias tables start at zero. \Cref{tab:optimization} consolidates training data and optimization settings for all stages.

Min--max statistics are computed per channel over valid training joints and shared across positions and robots. Features map to $[-1,1]$, with $10^{-6}$ added to the denominator. Padding is excluded. Raw angle radians are not wrapped. Decoder rotations are denormalized before conversion to matrices. \Cref{eq:tokloss} uses normalized angle/position errors and physical rotation geodesics divided by $\pi$.

The lazy loader reserves every 1000th global row for validation (fraction 0.001). This splits samples rather than morphology identities. The 40M count refers to morphology--state pairs. Unique pretraining URDF counts and sampling densities require the data manifest. Tokenizer training completes 300 epochs (365,700 optimization steps). Training uses eight GPUs, batch 4096 per GPU, and no gradient accumulation (global batch 32,768). All six joint-space policies and the EE reference complete 30,000 steps on RoboTwin~2.0 and 60,000 on EmbodiBench, with global batch 128.

\subsection{Ablation Configuration Differences}
Angle+M retains the joint-code layout, URDF branch, and graph biases but removes link-geometry prediction and supervision. Angle additionally removes the morphology branch, all three graph-bias types, and policy URDF conditioning. It also changes the commitment weight from 0.1 to 0.25. Thus Angle versus Angle+M changes several factors, while the full model versus Angle+M compares link-pose supervision under morphology conditioning. The code labels this output and loss ``EE,'' although they cover all valid links.

\subsection{Physical Reconstruction Metrics}
For each valid joint, angle error is the absolute decoder-to-target difference in degrees without periodic wrapping. Position error is the Euclidean distance in meters between the decoder's direct position output and FK of the original joint angles. Orientation error is the SO(3) geodesic angle in degrees between the decoded 6D rotation and the corresponding FK rotation. These metrics cover all valid child-link frames, not just the end effector. They do not evaluate FK of reconstructed angles.

Mean, P90, and maximum are computed after pooling all valid raw errors within each description group. This weights individual joint/link samples equally, without balancing trajectories, tasks, or morphologies. Longer trajectories and arms with more valid joints therefore contribute more samples. Policy success instead averages tasks equally, so the two metrics use different weighting schemes. Padding, missing arms, and grippers are excluded. Shardwise or jointwise percentiles are not averaged. The encoder receives joint angles and morphology, without link-pose input. Reconstruction excludes VLA code prediction and continuous refinement.

\subsection{Reconstruction Data Coverage}
\label{app:reconstruction-data}
We reconstruct every original saved frame of the downstream task datasets, rather than subsampled frames or the ten fitted action nodes. A time frame counts once regardless of camera count. Each real arm is reconstructed independently. RoboTwin~2.0 uses 750 trajectories per embodiment, totaling 3,750, consistent with the policy-training count. \Cref{tab:reconstruction-coverage} summarizes trajectory coverage.

EmbodiBench reconstruction uses the policy-training demonstrations from \cref{sec:embodibench}, with 1,500 trajectories per task and 15,000 in total. All downstream policy URDFs are excluded from tokenizer pretraining. EmbodiBench uses newly generated URDFs. This evaluates the frozen tokenizer on unseen morphologies, using demonstrations that also train the downstream policy. Each task uses 30 URDFs. The union of left/right arm IDs across tasks contains 55 distinct URDFs, also confirmed by SHA-256 deduplication of the actual URDF files. The dataset contains 1,850,860 original time frames, 3,151,136 real-arm frames, and 20,604,806 valid joint/link error samples. These 55 task-data morphologies are not the unique-morphology count of the 40M pretraining corpus.
% Author-confirmed trajectory counts, 2026-11-16. No inferred frame counts.
\begin{table}[t]
\centering
\small
\setlength{\tabcolsep}{12pt}
\caption{\textbf{Reconstruction evaluation coverage.} All original saved frames are evaluated. Each real arm is reconstructed independently. Padding, missing arms, and grippers are excluded. Generated URDFs span 55 distinct morphologies across ten task-specific sets of 30.}
\label{tab:reconstruction-coverage}
\begin{tabular}{@{}lr@{}}
\toprule
Description & Trajectories \\
\midrule
ALOHA & 750 \\
ARX-X5 & 750 \\
Franka Panda & 750 \\
Piper & 750 \\
UR5-WSG & 750 \\
\midrule
RoboTwin~2.0 total & 3,750 \\
Generated URDFs & 15,000 \\
\bottomrule
\end{tabular}
\end{table}

\section{VLA Training and Sampling Details}
\label{app:policy-settings}
\subsection{Observations, Action Layout, and Context}
Qwen3-VL-4B-Instruct receives $224\times224$ head/wrist RGB views and an instruction containing 256-bin current-state tokens. The action head also receives state directly. Each arm has eight slots, comprising seven arm joints and one gripper channel. Masks suppress absent joints and arms. Prediction and execution both use 50-frame chunks of absolute joint positions and gripper commands, passed to the simulator's controllers.

Ten fitted joint nodes per chunk (\cref{eq:temporalfit}), two IDs per joint, and one ID per gripper yield $10(2J+1)$ IDs per arm, or at most 300 for two seven-joint arms. Grippers use ordinary temporal resampling and 256 bins. The 256-bin joint baseline uses the same ridge fitting but clips fitted nodes to its fixed quantization interval. FAST uses variable-length autoregressive codes without node fitting. The pose head receives auxiliary geometric supervision during tokenizer pretraining. During control, the policy refines decoded joint actions and does not execute the predicted link poses.

LayerNorm and linear projection map each VLM layer's features to width 1024. The action head uses 16 heads, 32 future queries, position capacity 2048, dropout 0.2, and the VLM's layer count. Even blocks use VLM cross-attention. Odd blocks use action self-attention. The discrete and continuous branches share their first two blocks, with separate later blocks and no cross-branch attention. The VLM and action head are trainable. The entire tokenizer, including its morphology encoder, is frozen. Training uses BF16, FSDP, and AdamW epsilon $10^{-8}$.

\subsection{Discrete Training and Iterative Completion}
For $N$ valid token positions, training samples $u\sim\mathcal U(0,1)$ and masks
\begin{equation}
 n_{\mathrm{mask}}=\operatorname{clip}\!\left(\operatorname{round}\!\left[N\cos\left({\pi u\over2}\right)\right],1,N\right).
\end{equation}
Cross-entropy is averaged only over these masked valid positions. Inference begins with all valid positions masked and performs eight iterations. Token candidates are sampled at categorical temperature 1. For iteration progress $r\in(0,1]$, confidence combines the sampled token's log probability with Gumbel noise scaled by $1-r$. The next unknown count follows $\lfloor N\cos(\pi r/2)\rfloor$, capped below the current unknown count to ensure progress. Highest-confidence candidates are committed. Committed positions are never remasked.

\subsection{Continuous Source, Time Sampling, and Execution}
Each action coordinate maps its dataset's 1st/99th percentiles to $-1/1$, then clips to $[-2.2,2.2]$. Coarse and ground-truth actions share this normalization. Invalid coordinates are zero. Of eight repeated instances per demonstration, four train each branch. Masked CE and flow matching have weights 0.5 and 1. Optional angle and one-step losses are disabled.

The continuous source is $x_0=0.3a_d+0.7\epsilon$. To sample its training time, the implementation draws $b\sim\operatorname{Beta}(1.5,1)$ and sets
\begin{equation}
 t=\operatorname{clip}\!\left({0.999-0.5b\over0.999},0.5,1\right),\qquad \gamma=2t-1.
\end{equation}
The phase $t$ routes the continuous branch, while $\gamma$ controls interpolation in \cref{eq:flowloss} with velocity target $a^*-x_0$. Training reconstructs $a_d$ from ground-truth IDs. Rollout uses predicted IDs. Inference evaluates four Euler steps at $\gamma=0,0.25,0.5,0.75$, with step size 0.25, then denormalizes and executes all 50 frames. These iteration counts do not measure wall-clock latency.

\section{EmbodiBench Protocol Details}
\label{app:benchmark-details}
\subsection{Tasks, Robot Schedules, and Scene Seeds}
\Cref{tab:embodibench-tasks} lists the complete ten-task evaluation suite. Other implemented environments are excluded. Suffixes SS/OS denote same-side/opposite-side dual-arm placement. Both arms use the same selected morphology. Collection and evaluation follow task-indexed robot-ID schedules.

Evaluation requests 150 trials per task, five per URDF, with fixed instructions and expert feasibility checking. The initial scene seed is $100000(1+s)$ for evaluation seed $s$. Planner-infeasible scenes advance the seed stream without consuming a trial, so success rates are conditional on planner feasibility. Each accepted trial allows 300 executed commands. The counter increments per \texttt{take\_action}, including every chunk frame, rather than per policy query or physics step.

Default head/wrist cameras use $320\times240$ resolution and $45^\circ/37^\circ$ fields of view. The model resizes to $224\times224$. Scene randomization is set separately in the run configuration.

\Cref{fig:embodibench} uses uncropped keyframes from $640\times480$, 30-fps renders of successful saved trajectories (512 samples per pixel). Stacking frames show the released upper block. These are visualization replays, not learned-policy rollouts or policy observations.

% Keep the success definitions together on the final supplementary page.
\newpage
\subsection{Success Predicates}
\paragraph{Picking.} The relevant object centers must rise more than 0.07\,m from their initial heights.

\paragraph{Pushing.} The object must satisfy $|x|<0.03$ and $|y-0.05|<0.03$ meters. Its normalized quaternion rotation distance to the target must be below 0.05. The distance is $2\arccos(|\Delta q_w|)/\pi$, so the threshold corresponds to $9^\circ$.

\paragraph{Stacking.} The target is the lower block position plus $(0,0,0.05)$\,m, with axiswise tolerances $(0.025,0.025,0.012)$\,m for the upper block. The appropriate grippers must also open.

\paragraph{Lifting.} The reported results use an at-least-one-arm stable-grasp predicate. For 15 consecutive physics steps, the object center must satisfy $z>0.9$\,m and rise at least 0.03\,m above its settled initial height. At every step, both fingers of the same gripper must form a valid grasp and pass the gripper closure/holding check. Validity requires contact on the true inner finger surfaces, opposing grasp directions, and at least 0.5\,N effective normal force per finger, with surface tolerance 0.002\,m.

The qualifying arm may change during the interval, but contacts from different grippers cannot be combined into a grasp. Any failed condition resets the counter. At 250\,Hz, 15 physics steps correspond to 0.06\,s, not 15 policy actions. There is no object-tilt bound, support-separation requirement, or simultaneous two-arm grasp requirement. Each episode permits at most 300 action targets, predicted and executed in 50-step chunks.

% Put the result mapping and evidence scope in the facing column.
\newpage
\subsection{Result Mapping and Reproducibility Scope}
\label{app:tasks}
The implementation identifies methods by configuration names. Ours is \texttt{qwen\_w\_eat}. Angle+M is \texttt{qwen\_urdf\_angle}. Angle is \texttt{qwen\_angle\_only}. 256-bin and FAST are \texttt{qwen\_256bin} and \texttt{qwen\_fast}. Joint and EE denote the corresponding StarVLA control interfaces. Reported means are unweighted task averages. Recomputation from rounded entries differs by at most 0.011 percentage points. The tables report aggregate success rates without across-training-seed standard deviations.

Tokenizer pretraining and downstream policy learning use disjoint URDF sets. EmbodiBench reconstruction reuses policy-training demonstrations. It measures representation fidelity on morphologies unseen by the tokenizer, not policy generalization. No disjoint-morphology split is established between policy training and rollout evaluation. The 40M pretraining sample count does not specify the number of unique morphologies. Identity manifests are needed to reproduce the split. We do not report generation throughput, scaling curves, or variation across training seeds.

\end{document}